\documentclass[runningheads]{llncs}
\usepackage[T1]{fontenc}
\usepackage{booktabs}
\usepackage{graphicx}
\usepackage{hyperref}
\usepackage{subfig}
\usepackage{amsmath}

\begin{document}
\title{Sentence-to-Label Generation Framework for Multi-task Learning of Japanese Sentence Classification and Named Entity Recognition}

%
%

\author{Chengguang Gan\inst{1}\orcidID{0000-0001-8034-0993} \and
Qinghao Zhang\inst{2}\and
Tatsunori Mori\inst{1}\orcidID{0000-0003-0656-6518}}

\authorrunning{Gan et al.}

\institute{Yokohama National University, Japan\\
\email{gan-chengguan-pw@ynu.jp}, \email{tmori@ynu.ac.jp} \and
Department of Information Convergence Engineering, Pusan National University, South Korea\\
\email{zhangqinghao@pusan.ac.kr}}
\maketitle              

\begin{abstract}

Information extraction(IE) is a crucial subfield within natural language processing. In this study, we introduce a Sentence Classification and Named Entity Recognition Multi-task (SCNM) approach that combines Sentence Classification (SC) and Named Entity Recognition (NER). We develop a Sentence-to-Label Generation (SLG) framework for SCNM and construct a Wikipedia dataset containing both SC and NER. Using a format converter, we unify input formats and employ a generative model to generate SC-labels, NER-labels, and associated text segments. We propose a Constraint Mechanism (CM) to improve generated format accuracy. Our results show SC accuracy increased by 1.13 points and NER by 1.06 points in SCNM compared to standalone tasks, with CM raising format accuracy from 63.61 to 100. The findings indicate mutual reinforcement effects between SC and NER, and integration enhances both tasks' performance.

\end{abstract}
\section{Introduction}

\begin{figure*}[tp]
\centering
\includegraphics[width=347.25 pt]{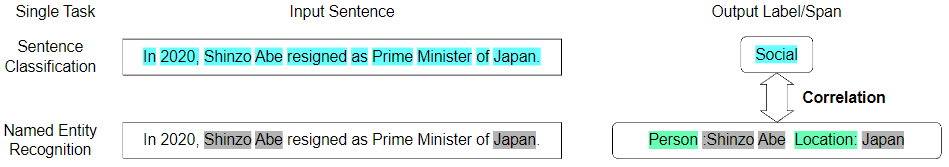}
\caption{\label{1figure1}The illustration depicts the interrelationship between the labels of Named Entity Recognition (NER) and Sentence Classification (SC) within a single sentence.}

\end{figure*}

In the realm of information extraction, numerous specialized tasks exist, such as named entity recognition\cite{ritter2011named}\cite{lample-etal-2016-neural}\cite{nadeau2007survey}, relation extraction\cite{mintz-etal-2009-distant}, event extraction\cite{etzioni2008open}, sentence classification\cite{zhang2015sensitivity}, sentiment analysis\cite{medhat2014sentiment}\cite{pang-etal-2002-thumbs}, and more. With the advent of the Transformer architecture, pre-training and fine-tuning paradigms have gained widespread adoption. Typically, models undergo unsupervised pre-training on a large-scale, general corpus, such as Wikipedia text, in order to acquire foundational knowledge. These pre-trained models are then fine-tuned for specific downstream tasks. However, due to the considerable variation in data features and requirements across tasks, adapting a single dataset and model to multiple tasks simultaneously is challenging. Consequently, researchers often create dedicated datasets for distinct IE tasks and employ these for fine-tuning pre-trained models. Moreover, IE methods have evolved from sequence labeling tasks utilizing Long Short-Term Memory (LSTM)\cite{huang2015bidirectional} to seq to seq generative IE methods\cite{lu-etal-2022-unified}. The emergence of generative approaches indicates the feasibility of addressing multiple tasks with a single model by unifying input and output formats. In the present study, illustrated by Figure \ref{1figure1}, the SC and NER tasks are fed as inputs to their respective fine-tuned models. Then models generate corresponding labels/spans for each task, respectively.


\par While generic pre-training knowledge can be beneficial for various downstream tasks, the possibility of mutual reinforcement effects between tasks remains an open question. To explore this, we hypothesize that mutual reinforcement effects exists between different tasks. To test this hypothesis, we focus on Named Entity Recognition(NER) and Sentence Classification(SC) as the most representative tasks in IE, as illustrated in Figure \ref{1figure1}. In the SC task, a model generates sentence classification labels given an input sentence. In the NER task, a model identifies entity spans in the input sentence and generates corresponding labels and spans. Many task scenarios require simultaneous sentence classification and entity extraction, but no existing dataset satisfies both requirements. Furthermore, SC and NER tasks exhibit correlations in the labels they extract. For instance, a sentence mentioning “Shinzo Abe” likely pertains to Social, and a Social sentence is more likely to contain names of Social figures and countries. Consequently, we investigate whether leveraging the interrelationship between SC and NER tasks can improve model performance.
\par In this context, we propose a novel framework for handling both Japanese SC and NER tasks. The primary contributions of this work include:
\par 1.Integrating SC and NER tasks into a new Sentence Classification and Named Entity Recognition Multi-task(SCNM) task and constructing an SCNM dataset by annotating SC labels on the existing Wikipedia Japanese NER dataset.
\par 2.Proposing a Sentence-to-Label Generation Framework(SLG) for addressing the SCNM task, comprising format converter construction, incremental training, and the development of a format Constraint Mechanism(CM). The format converter enables the SLG to handle the SCNM task, as well as SC or NER tasks separately, highlighting its generalizability.
\par 3.Demonstrating through ablation experiments that SC and NER tasks are mutual reinforcement effects. The performance of a model trained by combining both tasks surpasses that of a model fine-tuned on a single task, supporting the notion that 1+1>2. This finding offers insights for future scenarios requiring SCNM.
\par The remainder of this paper is structured as follows: Some related work and prior studies are presented in Section \ref{sec:related work}. In Section \ref{sec:Task Setup and Dataset Constructed}, we describe the task setup and the construction of the SCNM dataset used in our study. Section \ref{sec:Sentence-to-Label Generation Framework} presents the Sentence-to-Label Generation Framework, which encompasses the format converter \ref{sec:Format Converter}, Incremental Learning \ref{sec:Incremental Learning}, and Constraint Mechanism \ref{sec:Constraint Mechanism}. Lastly, Section \ref{sec:Experiments} discusses the results of our SLG framework experiments, as well as various ablation studies to evaluate the effectiveness of the proposed approach.

\section{Related Work}\label{sec:related work}
In this section, we provide a comprehensive overview of previous studies on generative IE methodologies. Furthermore, we delineate the similarities and distinctions between these prior works and our current research, thereby highlighting the novel contributions of our study.
\par \textbf{Word-level}. In \cite{yan-etal-2021-unified-generative}, the authors propose a novel Seq2Seq framework that addresses flat, nested, and discontinuous NER subtasks through entity span sequence generation. Similarly, \cite{chen-etal-2022-shot} used a self-describing mechanism for few-shot NER, which leverages mention describing and entity generation. GenIE\cite{josifoski-etal-2022-genie} uses the transformer model to extract unstructured text relationally through global structural constraint. And LightNER\cite{chen-etal-2022-lightner} is addresses class transfer by constructing a unified learnable verbalizer of entity categories and tackles domain transfer with a pluggable guidance module. InstructionNER\cite{wang2022instructionner} and UIE\cite{lu-etal-2022-unified} have also developed frameworks for word-level IE tasks.
\par \textbf{Sentence-level}. In terms of using sentences label to improve the NER effect. Joint learning framework used BiLSTM model and attention,CRF layer to  improve the effectiveness of NER through sentence labeling\cite{kruengkrai-etal-2020-improving}. MGADE that uses a dual-attention mechanism to concurrently address two Adverse Drug Event (ADE) tasks: ADE entity recognition at the word level (fine-grained) and ADE assertive sentence classification (coarse-grained). The model takes advantage of the interdependencies between these two levels of granularity to improve the performance of both tasks\cite{wunnava-etal-2020-dual}. 
\par In conclusion, this study employed a generative model for Sentence-to-Label conversion, distinguishing itself from previous studies by pioneering the utilization of mutual reinforcement effects between SC and NER tasks. This novel approach effectively enhanced the accuracy of both tasks. Additionally, we introduced the SLG framework, which enables the model to adeptly manage both SC and NER tasks simultaneously.

\section{Task Setup and Dataset Constructed}\label{sec:Task Setup and Dataset Constructed}

\begin{figure*}[!tp]
\centering
\includegraphics[width=347.25 pt]{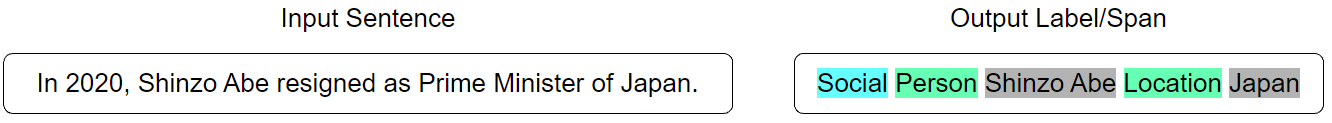}
\caption{\label{2evaluation}The illustration of Sentence Classification and Named Entity Recognition Multi-task(SCNM) task.}
\end{figure*}

Before delving into the SLG, let us first describe the structure of the SCNM task. SCNM involves the classification of both sentence-level and word-level information within a single sentence. As illustration of Figure \ref{2evaluation}, given an input sentence(i.e., In 2020, Shinzo Abe resigned as Prime Minister of Japan), the model is expected to generate a classification label(i.e., Social) for the sentence, along with named entity labels(i.e., Preson, Location) and the corresponding word spans(i.e., Shinzo Abe, Japan) present in the sentence. Moreover, when selecting and constructing the dataset, it is crucial to encompass a wide range of content, and the classification labels should be generic rather than specific to a narrow domain. Considering these factors, we chose the Japanese Wikipedia-based NER dataset\cite{omiya2021wikipedia} for our SCNM dataset foundation. It consists of 5,343 sentences (4,859 with named entities, 484 without) and includes 8 categories (13,185 total named entities): person, company, political org., other org., location, public facility, product, and event. This diverse and broad dataset is ideal for the SCNM task.
\par Following the selection of the NER dataset, we annotated sentence classification labels based on the original dataset. We partitioned the Wikipedia sentences into five primary categories: social, literature and art, academic, technical, and natural. All 5,343 sentences were categorized into these five groups, ultimately resulting in the construction of the SCNM dataset. Figure \ref{2evaluation} illustrates a specific instance of the input and output by the SCNM task.

\subsection{Evaluation}\label{Evaluation}

Upon constructing the SCNM dataset, the primary challenge we encounter is devising a comprehensive set of evaluation metrics for this new dataset. We categorize the SCNM evaluation metrics into two distinct classes: label evaluation and format evaluation.

\par
\begin{figure*}[!t]
\centering
\includegraphics[width=347.25 pt]{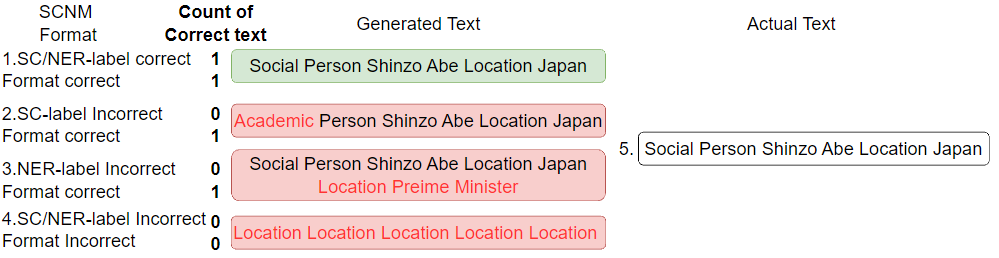}
\caption{\label{2_1evaluation}The illustration of text and format evaluation. 1 and 0 represents a correctly or incorrectly generated text, and the total number of all correctly generated text in the test set adds up to $C$.}

\end{figure*}

 \textbf{Text Evaluation}. As illustrated in Figure \ref{2_1evaluation}, we present several instances of both generated and actual text, labeled numerically (1-5) for further analysis. Given that the combined task of SC and NER precludes the computation of traditional metrics(e.g., precision recall f1-score) in the conventional manner, generative IE deviates from the traditional sequence tagging model. Specifically, in the generative IE approach, a token does not correspond to a label. Furthermore, the number of generated named entities and word spans is variable (e.g., 3), as is the number of generated text, which may also contain extraneous text (e.g., 4).

\par Consequently, we employ accuracy as the evaluation metric, when generate text 1 is all equal to actual text 5, it counts as a correct text 1. Although this evaluation criterion may seem somewhat stringent, it is well-suited for SCNM tasks. We denote generated text as $G$ and actual text as $A$. The $|A \cap G|$ is represented by $C_{\text{generated}\, \text{text}}$ (the total number of matches text between $G$ and $A$), while the total number of actual text(i.e, total number of samples in the test set) is denoted by $T_{\text{actual}\, \text{text}}$.

\begin{equation}
    \text{SCNM Accuracy} = \frac{{C}_{\text{generated}\, \text{text}}}{{T}_{\text{actual}\, \text{text}}}
\end{equation}

In addition, we also computed accuracy for SC and NER in the SCNM task separately. Specifically, the SC-label and all remaining NER-label/span in generated text and actual text were split. Then the accuracy is calculated separately.

\begin{equation}\label{scneraccuracy}
    \text{SC Accuracy} = \frac{C_{\text{SC}}}{T_{\text{SC}}} \qquad
    \text{NER Accuracy} = \frac{C_{\text{NER}}}{T_{\text{NER}}}
\end{equation}

\par \textbf{Format Evaluation}. In generative IE, a key metric is the ability to produce outputs in the right format. Due to the uncontrollable nature of generative models, there's a high chance of generating duplicate words or incoherent sentences. Incorrect output formats likely lead to wrong label classification (e.g., 4). Thus, we augment label evaluation with format generation accuracy assessment.
\par This additional evaluation aims to gauge a model's proficiency in controlling the generated format. If the first generated text becomes SC-label and the subsequent generated ones are NER-label and span, it is counted as a format correct number (regardless of whether the generated SC-label and NER-label are correct or not, they are counted as format correct). As illustrated in Figure \ref{2_1evaluation} (e.g., 1 2 3). The total number of generated text with correct format is $C_{\text{format}}$, and the total number of actual text is $T_{\text{format}}$. The Format Accuracy is defined as:

\begin{equation}
    \text{Format Accuracy} = \frac{C_{\text{format}}}{T_{\text{format}}}
\end{equation}

\begin{figure*}[htp]
\centering
\includegraphics[width=347.25 pt]{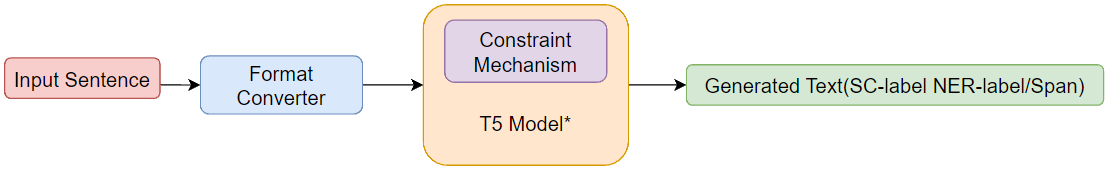}
\caption{\label{3pipline}The illustration of overview for Sentence-to-Label Generate(SLG) Framework.  The T5 model* is the vanilla T5 model with incremental learned using Shinra NER corpus.}

\end{figure*}

\section{Sentence-to-Label Generation Framework}\label{sec:Sentence-to-Label Generation Framework}
In the preceding section, the SCNM task setting and dataset construction were presented. This section offers a thorough overview of the Sentence-to-Label (SL) framework, followed by a detailed explanation of each component of the SLG in three separate subsections.
\par An overview of the SLG framework is depicted in Figure \ref{3pipline}. Initially, the Shinra NER corpus\footnote{\raggedright\url{http://shinra-project.info/shinra2020jp/data_download/}} is reformatted using a format converter. Subsequently, the transformed corpus serves as a basis for incremental learning in the model. The SCNM dataset, converted by the format converter, is then fine-tuned for the model. Prior to the model's Decoder generating prediction results, a Constraint Mechanism(CM) is incorporated to enhance the model's format generation capabilities. Lastly, the SC-label, NER-label, and corresponding word span are sequentially outputted.

\subsection{Format Converter}\label{sec:Format Converter}

In this subsection, we explore multiple converter formats and evaluate performance in later experiment sections. The most effective format is chosen for our converter.

\begin{figure*}[!h]
\centering
\includegraphics[width=347.25 pt]{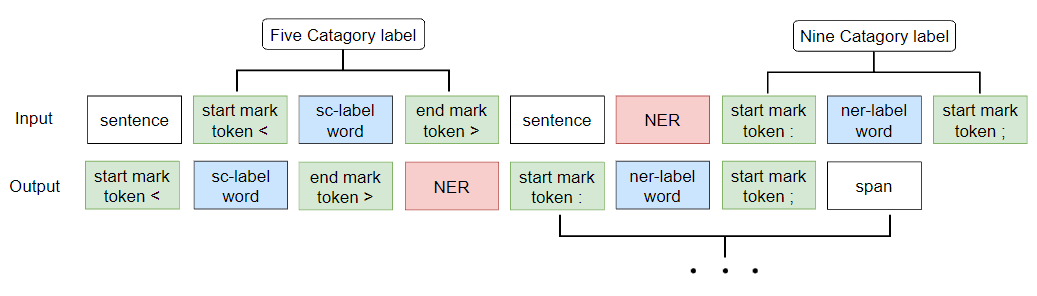}
\caption{\label{4formatconverter}The illustration of format converter.}
\end{figure*}

\par As show in Figure \ref{4formatconverter}, the optimal format is determined from the experimental outcomes. For the input sequence of the model, the original sentence is positioned at the beginning, succeeded by five SC-label words. The start and end tokens, “<” and “>”, respectively, enclose all SC-labels. Subsequently, the original sentence is repeated immediately after the above end mark token“>”. To signal the model's initiation of NER-label generation and corresponding word span, a prompt word “NER” is appended after the sentence.
\par Due to the presence of negative sentences in the SCNM dataset that lack named entities, an additional “None” label is introduced for negative cases, augmenting the original eight NER-labels. Consequently, a total of nine NER-labeled words follow the prompt word. To indicate the commencement and termination of the NER-label word, the start and end tokens, “:”, and “;”, are employed, respectively. The distinct mark tokens for SC and NER labels demonstrate superior performance in the experiments, as compared to identical mark tokens.
\par
\begin{figure*}[htp]
\centering
\includegraphics[width=347.25 pt]{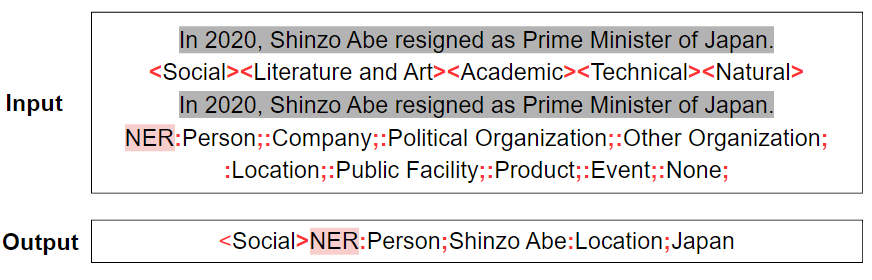}
\caption{\label{5formatconvertersample}A specific example of an input converted using the format converter, and the corresponding generated output.}
\end{figure*}

 Regarding the output format of the model, the overall and input token order is maintained consistently. However, it is crucial to note that only one of the predicted SC-label words is utilized, rather than all five words present in the input. Following the starting word “NER”, the NER-label word and corresponding word span derived from the sentence are provided. Utilizing the format converter, the model demonstrates versatility in managing diverse tasks, such as SC, NER, and SCNM. These tasks can be effectively addressed either individually or concurrently. And through format converter, the model learns the correct format to generate a uniform format. A specific example of the input and output sequence is show in Figure \ref{5formatconvertersample}.

\subsection{Incremental Learning}\label{sec:Incremental Learning}

In this subsection, we elucidate the process of applying incremental learning(IL) to the vanilla T5 model\footnote{\raggedright\url{https://huggingface.co/sonoisa/t5-base-japanese}} using the Shinra NER corpus, as well as the underlying rationale. The T5 model is primarily pre-trained for sequence-to-sequence (seq2seq) tasks in text processing\cite{raffel2020exploring}. However, it lacks specific pre-training for word-level attributes, such as named entities. Our objective is to implement IL in a seq2seq format, tailored for named entities, without causing the model to lose its pre-trained knowledge.

\par
\begin{figure*}[t]
\centering
\includegraphics[width=347.25 pt]{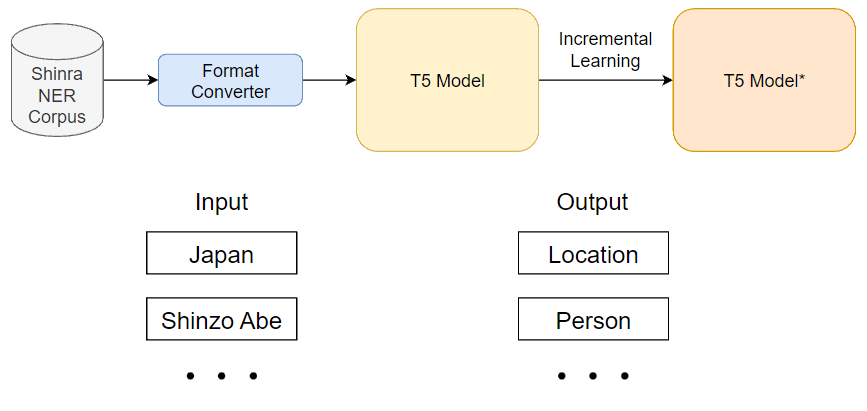}
\caption{\label{7incrementallearning}The illustration of incremental learning. And IL specific format converter for Shira NER corpus.}
\end{figure*}

 We selected the Shinra2020-JP corpus as the data source for IL. The Shinra project, an extended named entity recognition (NER) endeavor, is constructed based on the Japanese Wikipedia. Consequently, the corpus encompasses a wide array of named entities derived from Wikipedia. In other words, this corpus contains various named entities of wikipedia. The categories of named entities are: facility, event, organization, location, airport name, city, company, Compound, person. It is a relatively comprehensive and extensive NER corpus. The dataset is used for incremental learning of the T5 model after a simple format transformation(e.g., input:Japan, output:Location). The total number of samples is 14117 and there are no duplicate samples. As illustrated in the figure \ref{7incrementallearning}, the T5 model employed within the SLG framework is ultimately acquired.

\subsection{Constraint Mechanism}\label{sec:Constraint Mechanism}

To enhance the accuracy of the model's output format, we introduce an efficient and straightforward Constraint Mechanism (CM). Let $X_1$ denote the initial token predicted by the Decoder, with the total predicted sequence comprising $n$ tokens. The predicted sequence can be represented as $(X_1, X_2, \dots, X_n)$. The prediction probability for the second token $X_2$ can be formulated as the subsequent equation:
\begin{equation}
    P(X_2 | X_1) = {Decoder_{2}}(X_1,Encoder(Inputs))
\end{equation}
Here, $Encoder(Inputs)$ denotes the vector resulting from the computation of the input by the next layer of Encoder. $Decoder_{2}$ denotes the result of the computation of the vector output by encoder and the first token vector output by Decoder of the first layer is passed to the second layer Deocder.

\begin{figure*}[!t]
\centering
\includegraphics[width=347.25 pt]{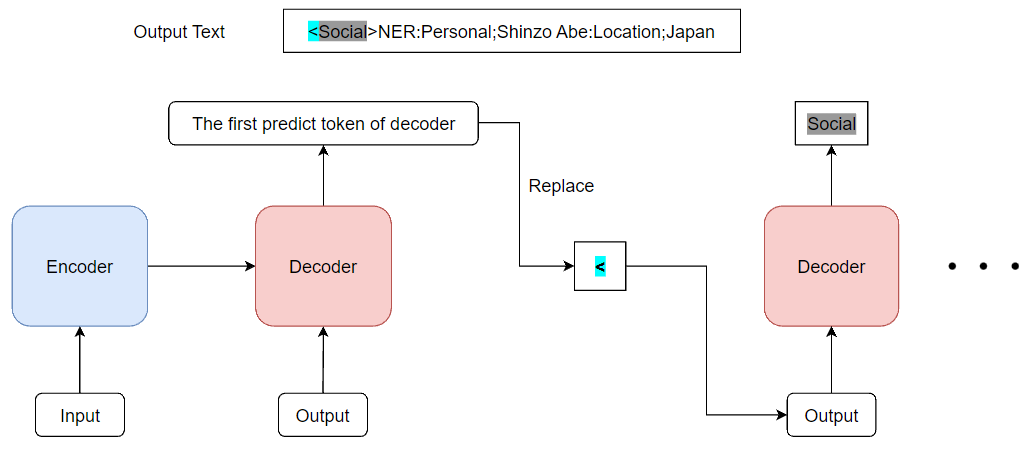}
\caption{\label{6constraintmechanism}The processes of Constraint Mechanism(CM).}
\end{figure*}

In Figure \ref{6constraintmechanism}, a specific example is presented. The output text refers to the desired output sequence that should be generated accurately. Notably, the first token of every output text within the SCNM dataset remains constant, represented by the “<” symbol. Consequently, the initial token of each predicted sequence is compelled to be replaced with the “<” token. In other words, this substitution corresponds to the numerical value of the “<” symbol in the T5 model's vocabulary.
\par Adopting this approach ensures that the first predicted token in each newly generated output text is accurate, which in turn enhances the precision of subsequent token predictions. The ultimate experimental outcomes corroborate that the CM technique effectively augments the model's capacity to generate accurate formats, thereby improving the overall correctness of SCNM tasks.

\section{Experiments}\label{sec:Experiments}

In this chapter, we present a comprehensive series of experiments conducted on our proposed SLG framework and its application to SCNM tasks. Our aim is to demonstrate the effectiveness of the SLG framework and the synergistic effect of integrating the SC and NER tasks.
\par Given that the Japanese T5 model is only available in base size, we employ the T5-base as the underlying model weight for the SLG framework. To ensure robustness, we conduct each experiment three times, and the final result is obtained by averaging the outcomes. The proportion used for dividing the data into the training set and the test set is 9:1. We adopt a randomized approach to data set partitioning, using the parameter “Random Seed=None” to guarantee distinct training and testing sets for each iteration.  For an in-depth discussion of the evaluation metrics utilized, please refer to Section \ref{Evaluation}.

\subsection{Result on SLG with SCNM task}

\begin{table*}[!t]
\centering
\setlength{\tabcolsep}{0.25cm}
\caption{\label{table1}
The result of SLG framework with SCNM dataset. Accuracy was used for all evaluation metrics. SCNM represents SCNM task. SCNM* represents SCNM dataset.
}
\begin{tabular}{cccc}
\toprule[2pt]
  \textbf{Dataset} & \textbf{SCNM Accuracy} & \textbf{SC Accuracy} & \textbf{NER Accuracy}\\
\midrule
\textbf{SCNM*} & \textbf{72.41} & \textbf{88.89} & \textbf{81.96} \\
\textbf{SC Only} & — & 87.76 & — \\ 
\textbf{NER Only} & — & — & 80.90 \\
\bottomrule[2pt]
\end{tabular}
\end{table*}

In this subsection, we conduct a comprehensive evaluation experiment utilizing the SLG framework. Table \ref{table1} presents the results, with the first row displaying the metrics SCNM, SC, and NER, which correspond to the overall accuracy of the SCNM dataset, the accuracy of the SC dataset individually, and the accuracy of the NER dataset individually.
\par Three distinct datasets are outlined in the first column. “SCNM*” represents the full SCNM dataset, encompassing both SC and NER tasks. “SC Only” and “NER Only” signify the evaluation of the SC and NER datasets independently. To achieve this, we employ a format converter to separate the SC and NER components within the SCNM dataset, resulting in two distinct datasets (i.e., SC dataset and NER dataset). All three datasets are assessed using the SLG framework.
\par The evaluation reveals that the SCNM task attains a notable score of 72.41, even under rigorous evaluation metrics. Furthermore, by integrating the SC and NER tasks, the accuracy of NER improves by 1.06 compared to evaluating the NER task individually. Similarly, the SC task accuracy increased by 1.13 compared to evaluating it separately. The experimental outcomes highlight the exceptional performance of the SLG framework in handling the SCNM task, as well as the mutual reinforcement effects observed between the SC and NER tasks.

\subsection{Compare the effect of different formats on results}

In order to investigate the impact of various formats on the outcomes, we compared four distinct formats. Table \ref{table2} displays these formats, where the first row represents the input text and the second row signifies the output text. Given that the number of NER-label span pairs generated in each sentence is indefinite, we use *x to denote the count of generated NER-label span pairs. Moreover, since there are five SC-labels in the input text, we represent this with *5. The corresponding NER-labels total nine, denoted by *9, which includes eight labels from the original dataset and one additional “None” label.
\par In the experimental setup, we aimed to minimize the influence of external factors on the results. Therefore, only the original T5-base model was employed in the format comparison experiment, with the random seed and other hyperparameters held constant.
\par Table \ref{table2} reveals that the accuracy of the simplest format is considerably low on the SCNM dataset, particularly when the input consists solely of sentences. In this case, the accuracy is 0, with all results being incorrect. This is due to the model's inability to generate the desired format accurately. As a result, even if the SC or NER labels are generated correctly, they are still considered errors based on the strict evaluation criteria. The second format introduced a simple prompt word, “sentence NER”, which slightly improved accuracy to 0.19; however, the accuracy remained substantially low.
\par In the third and fourth formats, we incorporated all the SC-labels (five label words) and NER-labels (nine label words) into the input text. Upon adding sufficient prompt words to the input text, the accuracy of the third format increased to 0.37 and 0.56 respectively. Lastly, we modified the start and end mark tokens of the SC-label to “<” and “>”, respectively, to facilitate the model's differentiation between SC-label and NER-label. Consequently, the accuracy significantly improved to 29.40, surpassing the results of all previous formats. This format comparison experiment highlights the critical role of prompt design in obtaining accurate outcomes.

\begin{table*}[!t]
\centering
\caption{\label{table2}
The result of compared different format with SCNM tasks. And without IL and CM. The first line is input text, and second line is output text. 
}
\begin{tabular}{cl}
\toprule[2pt]
 \textbf{Different Format} & \textbf{Accuracy}\\
\midrule
 $\{$sentence$\}$ &  \\
 :SC-label;(:NER-label;span)*x & 0 \\
 
\midrule
 sentence:$\{$sentence$\}$ &  \\
 label:SC-label;NER(:NER-label;span)*x & 0.19 \\
 
 \midrule
 $\{$sentence$\}$category(:SC-label;)*5$\{$sentence$\}$NER(:NER-label;)*9 &  \\
 category:SC-label;NER(:NER-label;span)*x & 0.37 \\
 
\midrule
 $\{$sentence$\}$(\textbf{:}SC-label\textbf{;})*5$\{$sentence$\}$NER(:NER-label;)*9 &  \\
 \textbf{:}SC-label\textbf{;}NER(:NER-label;span)*x & 0.56 \\
 
\midrule
$\{$sentence$\}$(\textbf{<}SC-label\textbf{>})*5$\{$sentence$\}$NER(:NER-label;)*9 &  \\
 (\textbf{<}SC-label\textbf{>}NER:NER-label;span)*x & \textbf{29.40} \\
 
\bottomrule[2pt]
\end{tabular}
\end{table*}

\subsection{Ablation Study}

To assess the impact of SL and CM on the outcomes within the SLG framework, we carried out a series of ablation experiments. Table \ref{table3} illustrates that, upon removing Incremental Learning, the Named Entity Recognition NER accuracy within the SCNM dataset experienced a substantial decline of 15.98. Concurrently, the SC in the SCNM dataset also saw a reduction of 3.18. These findings highlight that incremental training not only enhances NER in the SCNM dataset but also exerts a positive influence on SC. This observation aligns with our initial hypothesis that SC and NER tasks exhibit a mutual reinforcement effects.

Conversely, when CM was eliminated from the SLG framework, the format accuracy of the model plummeted to 63.61. Consequently, the SCNM, SC, and NER metrics also witnessed significant decreases, falling to 46.38, 55.99, and 52.62, respectively. These experimental outcomes underscore the efficacy of our proposed CM approach. By effectively managing the initial token generated by the model, we can guide the subsequent tokens towards accurate generation, thereby improving the model's overall performance.

\begin{table*}[!h]
\centering
\setlength{\tabcolsep}{0.05cm}
\caption{\label{table3}
The result of IL or CM with ablation experiment. 
}
\begin{tabular}{ccccc}
\toprule[2pt]
  \textbf{Method} & \textbf{SCNM Accuracy} & \textbf{SC Accuracy} & \textbf{NER Accuracy} & \textbf{Format Accuracy}\\
\midrule
\textbf{SLG} & \textbf{72.41}  & \textbf{88.89} & \textbf{81.96} &  \textbf{100} \\
\textbf{ w/o IL} & 56.18 & 85.71 & 65.98 & 99.50  \\
\textbf{ w/o CM} & 46.38 & 55.99 & 52.62 & 63.61  \\
\bottomrule[2pt]
\end{tabular}
\end{table*}

\section{Conclusion and Future Work}

In this study, we integrate Sentence Classification (SC) and Named Entity Recognition (NER) tasks, leveraging their shared knowledge to enhance accuracy.  We propose the SCNM task and construct a comprehensive dataset from Wikipedia.  Our experiments demonstrate mutual reinforcement effects between SC and NER, and introduce the versatile Sentence-to-Label Generate (SLG) framework for handling both tasks concurrently or individually through a Format Converter.
\par Future work includes exploring alternative language models, assessing the SLG framework on other SC and NER datasets, and creating domain-specific SCNM datasets to evaluate the framework's adaptability and effectiveness.

\subsubsection{Acknowledgements} This research was  partially supported by JSPS KAKENHI Grant Numbers JP23H00491 and JP22K00502.

\label{sec:bibtex}

\bibliographystyle{splncs04}
\bibliography{References}



%
%
%

%

\end{document}